\documentclass[10pt,onecolumn,letterpaper]{article}

\usepackage{cvpr}
\usepackage{times}
\usepackage{epsfig}
\usepackage{graphicx}
\usepackage{amsmath}
\usepackage{amssymb}

\usepackage{algorithm}
\usepackage{algorithmic}
\usepackage{subfigure}

\DeclareMathOperator{\softmax}{softmax}
\DeclareMathOperator{\sigmoid}{sigmoid}
\DeclareMathOperator{\MMD}{MMD}

\cvprfinalcopy 

\def\cvprPaperID{2012} 

\ifcvprfinal\fi
\begin{document}

\title{Deep Transfer Network: Unsupervised Domain Adaptation}

\author{Xu~Zhang\\
{\tt\small spongezhang@gmail.com}\\
Shengjin Wang\\
{\tt\small wgsgj@tsinghua.edu.cn}\\
Tsinghua University
\and
Felix Xinnan Yu\\
{\tt\small yuxinnan@ee.columbia.edu}\\
Shih-Fu Chang\\
{\tt\small sfchang@ee.columbia.edu}\\
Columbia University
}

\maketitle

\begin{abstract}
Domain adaptation aims at training a classifier in one dataset and applying it to a related but not identical dataset.
One successfully used framework of domain adaptation is to learn a transformation to match both the distribution of the features (marginal distribution), and the distribution of the labels given features (conditional distribution).
In this paper, we propose a new domain adaptation framework named Deep Transfer Network (DTN), where the highly flexible deep neural networks are used to implement such a distribution matching process.

This is achieved by two types of layers in DTN: the shared feature extraction layers which learn a shared feature subspace in which the marginal distributions of the source and the target samples are drawn close, and the discrimination layers which match conditional distributions by classifier transduction.
We also show that DTN has a computation complexity linear to the number of training samples, making it suitable to large-scale problems.
By combining the best paradigms in both worlds (deep neural networks in recognition, and matching marginal and conditional distributions in domain adaptation), we demonstrate by extensive experiments that DTN improves significantly over former methods in both execution time and classification accuracy.
\end{abstract}

\section{Introduction}

Conventional machine learning requires a large amount of training samples in order to train a reliable model. In many real-world applications, however, it's expensive and sometimes even impossible to get enough labeled training data. A straightforward solution is to train a model on a labeled dataset, related but not identical to the target task, and apply the model to the data being considered. Unfortunately, performance of the model may drop dramatically during the domain transfer, due to different feature and label distributions \cite{dollar2009pedestrian}. In order to solve this problem, domain adaptation which aims at learning from one dataset (source dataset) and transferring the knowledge to a related but not identical dataset (target dataset) becomes an active research topic \cite{pan2010survey}. Such methods are being widely used in object classification \cite{bergamo2010exploiting}, object detection \cite{donahue2013semi,guillaumin2012large} one-shot learning \cite{fei2006one} \etc.

The key of most domain adaptation methods is to learn a transformation on the features to reduce the discrepancy of the distributions between the source and the target datasets. There are different situations in real-world problems. 1) The marginal distributions are different, while the conditional distributions are similar. 2) The marginal distributions are similar, while the conditional distributions are different. 3) Both the marginal and the conditional distributions are different \cite{zhang2013domain}.

Instance reweighting \cite{bruzzone2010domain,chu2013selective} and subspace learning \cite{gong2012geodesic,gong2013connecting,Long_2013_CVPR,quanz2012knowledge,pan2011domain} are two typical learning strategies for domain adaptation \cite{long2014transfer}. The former reduces the distribution discrepancy by reweighting the source samples, and trains a classifier on the weighted source samples. The latter tries to find a shared feature space in which the distributions of the two datasets are matched. There are also methods performing instance reweighting and subspace learning simultaneously and achieving the state-of-the-art results in many benchmark datasets \cite{long2014transfer}.

Classifier transduction represents another independent line of research for domain adaptation \cite{long2014adaptation}. It directly designs an adaptive classifier by incorporating distribution adaptation with model regularization \cite{duan2012domain,mirrashed2013domain,long2014adaptation}. A typical strategy is to use certain regularization on the conditional distribution in the optimization process. For example, minimizing a metric between the outputs of a classifier can reduce the discrepancy of the conditional distributions of the source and the target datasets \cite{long2014adaptation}, thus an adaptive classifier can be trained.

Recently, it has been shown in many challenging cases, both the marginal and the conditional distributions may be different between different domains. Simultaneously matching both of them can significantly improve the performance of the final classifier \cite{zhang2013domain,long2014transfer,long2014adaptation}. Methods belonging to this category achieved outstanding performance on benchmark datasets. However, many of them have high computational complexity ($\mathcal{O}(n^2)$ in \cite{long2014transfer} and $\mathcal{O}(n^3)$ in \cite{long2014adaptation}). It makes computations prohibitively expensive on large-scale problems.

Neural networks based deep learning approaches have achieved many inspiring results in machine learning and pattern recognition \cite{krizhevsky2012imagenet}. Recently, neural networks have been successfully used in solving domain adaptation such as sentiment classification \cite{glorot2011domain} and pedestrian detection \cite{zeng2014deep}. However there are few works solving general domain adaptation problem by explicitly matching the distributions of datasets through neural networks. Such explicit matching strategy has been shown to be crucial in the state-of-the-art methods \cite{zhang2013domain,long2014transfer,long2014adaptation}.

In this work, we propose the Deep Transfer Network (DTN), where a deep neural network is used to model and match both the marginal and the conditional distributions. The marriage of deep neural networks and domain adaptation provides us three unique advantages.
\begin{itemize}
\setlength{\itemsep}{-0.2\itemsep}
\item The neural network structure makes our model suitable to achieve domain transfer by simultaneously matching both the marginal and the conditional distributions.
We achieve this by two different types of layers in DTN: the shared feature extraction layer which learns a subspace to match the marginal distributions of the source and the target samples, and the discrimination layer which matches the conditional distributions by classifier transduction (Section \ref{sec:dtn}).

\item Compared to the former works, the proposed method can be optimized in $\mathcal{O}(n)$ time, where $n$ is the number of training data points. Compared to other methods ($\mathcal{O}(n^2)$ in \cite{long2013transfer,long2014transfer} and $\mathcal{O}(n^3)$ in \cite{long2014adaptation}), it is more suitable to deal with large-scale domain adaptation problems (Section \ref{sec:opt} and \ref{sec:exp}).

\item By combining the best paradigms of both worlds (neural networks in recognition, and matching both the marginal and the conditional distributions in domain adaptation),
we show by extensive experiments that the proposed method outperforms the state-of-the-art methods in both computation time, and accuracy.
In order to show our method on a large-scale setting, we have tested on two datasets, which are 10 times the size of the widely used Office-Caltech dataset.
Impressively, on USPS/MNIST dataset, DTN achieves 28.95\% improvement in accuracy (Section \ref{sec:exp}).
\end{itemize}

\section{Deep Transfer Network}
\label{sec:dtn}
\subsection{Problem Formulation}
Let $\mathbf{x}\in \mathbb{R}^d $ be a $d$-dimension column vector. $\mathbf{x}^s \in \mathbb{R}^d$ and  $\mathbf{x}^t \in \mathbb{R}^d $ represent samples in the source and the target datasets respectively. $y \in \mathbb{R}$ is the corresponding label of $\mathbf{x}$. Given a labeled source dataset $D^s = \{(\mathbf{x}^s_1,y^s_1),\ldots,(\mathbf{x}^s_{n^s},y^s_{n^s})\}$ and an unlabeled target dataset $D^t = \{\mathbf{x}^t_1,\ldots,\mathbf{x}^t_{n^t}\}$, where $n^s$ and $n^t$ are the numbers of the data points, the goal of domain adaptation is to learn a statistical model using all the given data to minimize the prediction error $\sum_{i = 1}^{n^t}\parallel \bar{y}^t_i - y^t_i \parallel$ in the target dataset, where $\bar{y}^t_i$ is the predected label of the $i~$th target data point by the model, and $y^t_i$ is the corresponding true label, unknown in training. We consider the case where both the marginal distributions and the conditional distributions of the source and the target datasets are different. In the following of this paper, $P(\mathbf{x}^s)$ and $P(\mathbf{x}^t)$ represent the marginal distributions of the source and the target datasets, while $P(y^s|\mathbf{x}^s)$ and $P(y^t|\mathbf{x}^t)$ represent the conditional distributions.

\subsection{Preliminary}
In DTN, distribution matching strategy is applied. In order to match the distributions, a metric of difference between two distributions needs to be defined. We follow \cite{Long_2013_CVPR} and adopt the empirical Maximum Mean Discrepancy (MMD) as the nonparametric metric.

MMD is a commonly used metric of discrepancy between two distributions, due to its
efficiency in computation and optimization \cite{quadrianto2009distribution}.
Denote $\mathbf{X}^s = [\mathbf{x}^s_1,\ldots,\mathbf{x}^s_{n^s}] \in \mathbb{R}^{d \times n^s}$ and $\mathbf{X}^t = [\mathbf{x}^t_1,\ldots,\mathbf{x}^t_{n^t}] \in \mathbb{R}^{d \times n^t}$ as the data matrices of $D^s$ and $D^t$ respectively. Denote the data matrix $\mathbf{X} = [\mathbf{X}^s,\mathbf{X}^t]$ as the combination of $\mathbf{X}^s$ and $\mathbf{X}^t$. Then MMD between the marginal distributions of the source and the target datasets is defined as:
\begin{equation}
\label{eq:MMD}
\begin{aligned}
\MMD &= \parallel \frac{1}{n^s}\sum_{i = 1}^{n^s}\mathbf{x}^s_i- \frac{1}{n^t}\sum_{j = 1}^{n^t}\mathbf{x}^t_j \parallel_2^2\\
               &=  \text{Tr}(\mathbf{X}\mathbf{M}\mathbf{X}^T),
\end{aligned}
\end{equation}
where $\mathbf{M}$ is the MMD matrix. Let $M_{ij}$ be one element of $\mathbf{M}$. $M_{ij}$ can be calculated as:
\begin{equation}
\label{eq:MMDMatrix}
M_{ij} = \left\{
        \begin{array}{ll}
        1/(n^s)^2, & i \leqslant n^s, j \leqslant n^s \\
        1/(n^t)^2, & i > n^s, j > n^s \\
        -1/(n^s n^t), & \textrm{otherwise}
        \end{array}
        \right. .
\end{equation}

A typical neural network consists of two types of layers:
\begin{itemize}
\setlength{\itemsep}{-0.2\itemsep}
\item Feature extraction layer, which projects the input data onto another space by linear projection (or convolution) followed by an nonlinear activation function.
\item Discrimination layer, where the final label prediction is preformed.
\end{itemize}

\begin{figure}
\centering
    \includegraphics[width=0.6\linewidth]{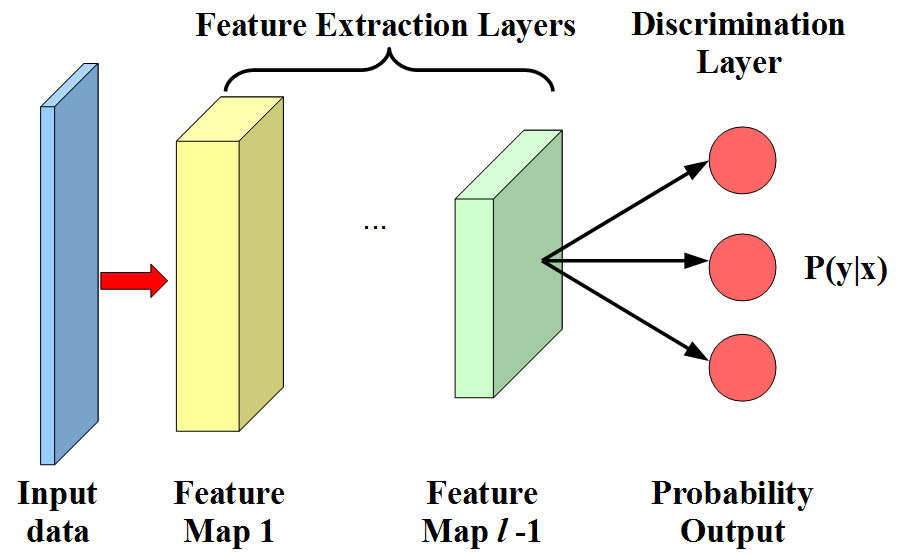}
   \caption{An example of MLP with $l$ layers. The first $l-1$ layers are feature extraction layers, and the last layer is the discrimination layer.}
   \label{fig:MLP}
\end{figure}

In the Convolutional Neural Networks (CNNs), for example, the feature extraction layers consist of the convolutional and fully connected layers, and the discrimination layer consist of the softmax regression layer.
In the Multi-Layer Perceptrons (MLPs) with $l$ layers (shown in Figure \ref{fig:MLP}), the first $l-1$ fully connected layers are feature extraction layers, while the last softmax regression layer is the discrimination layer.
In the following, we will develop DTN based on MLPs. Our method can be easily extended to other types of neural networks.

\subsection{Matching Marginal Distributions}
MLPs define a family of functions. We consider a single layer neural network first. The single layer neural network maps $\mathbf{x}$ to an k-dimension feature vector $\mathbf{h}$ by a linear projection and a nonlinear vector-valued activation function $f(\cdot)$:
\begin{equation}
\label{eq:single_layer}
\mathbf{h} = f(\mathbf{W}\mathbf{x}),
\end{equation}
where $\mathbf{W}$ is an $k \times d$ projection matrix. The typical choices of $f(\cdot)$ include $\tanh(a) = \frac{(e^a-e^{-a})}{(e^a+e^{-a})}$ and $\sigmoid(a) = \frac{1}{(1+e^{-a})}$.
The single layer neural network can be connected to form a deep neural network, where the output of one layer is used as the input of another layer.

For a MLP with $l$ layers, the first $l-1$ layers are all feature extraction layers. Let $\mathbf{h}(l-1)$ be the feature of $\mathbf{x}$ in the $l-1$th layer. Denote $P(\mathbf{h}^s(l-1))$ and $P(\mathbf{h}^t(l-1))$ as the distributions of the features of the source and the target datasets.
The goal of our method is to make $P(\mathbf{h}^s(l-1))$ and $P(\mathbf{h}^t(l-1))$ to be close.
This is achieved by minimizing the MMD when training the feature mapping function.

Let $\mathbf{H}^s(l-1) = [\mathbf{h}^s_1(l-1),\ldots,\mathbf{h}^s_{n^s}(l-1)] \in \mathbb{R}^{k \times n^s}$ be the feature matrix of the source dataset, while $\mathbf{H}^t(l-1) = [\mathbf{h}^t_1(l-1),\ldots,\mathbf{h}^t_{n^t}(l-1)] \in \mathbb{R}^{k \times n^t}$ be the feature matrix of the target dataset. And $\mathbf{H}(l-1) = [\mathbf{H}^s(l-1),\mathbf{H}^t(l-1)]$ is the combination of $\mathbf{H}^s(l-1)$ and $\mathbf{H}^t(l-1)$. The distance between $P(\mathbf{h}^s(l-1))$ and $P(\mathbf{h}^t(l-1))$ is modelled by the marginal MMD as follows:
\begin{equation}
\label{eq:MMD_Feature}
\begin{aligned}
\MMD_{mar}
 &= \parallel \frac{1}{n^s}\sum_{i = 1}^{n^s}\mathbf{h}^s_i(l-1)- \frac{1}{n^t}\sum_{j = 1}^{n^t}\mathbf{h}^t_j(l-1) \parallel_2^2\\
 &= \text{Tr}(\mathbf{H}(l-1)\mathbf{M}\mathbf{H}^T(l-1)),
\end{aligned}
\end{equation}
where $\mathbf{M}$ is the MMD matrix defined in Eq. \ref{eq:MMDMatrix}. Later, we will show how to minimize the above.

\subsection{Matching Conditional Distributions}
In this paper, we consider the logistic regression as the model of the discrimination layer.
It projects the input data points to a set of hyperplanes, and the distances to the planes reflect the posterior probabilities. Assume there are a total of $C$ categories in the dataset. For an arbitrary category $c$, the hyperplane of category $c$ is denoted as $\mathbf{w}_c$. The posterior probability of $y = c$ given feature $\mathbf{x}$ can be modelled as:
\begin{equation}
\label{eq:logistic_Regression}
P(y=c|\mathbf{x}) = \softmax_c(\mathbf{w}^T_c\mathbf{x}) = \frac {e^{\mathbf{w}^T_c\mathbf{x}}} {\sum_j e^{\mathbf{w}^T_j\mathbf{x}}}.
\end{equation}

The distance between the conditional distributions of the source and the target dataset is measured by the conditional MMD, defined as:
\begin{align}
\label{eq:MMD_Conditional}
\nonumber
&\MMD_{con}\\
\nonumber
&= \sum_{c = 1}^C \parallel \frac{1}{n^s}\sum_{i = 1}^{n^s}P(y^s=c|\mathbf{x}^s_i)- \frac{1}{n^t}\sum_{j = 1}^{n^t}P(y^t=c|\mathbf{x}^t_j) \parallel^2\\
&= \sum_{c = 1}^C \mathbf{q}_{c}^T\mathbf{M}\mathbf{q}_{c},
\end{align}
where $\mathbf{q}_{c} \in \mathbb{R}^{n^s+n^t}$ is the posterior distribution output vector of the $c~$th category for all the data points, and $\mathbf{M}$ is the MMD matrix defined in Eq. \ref{eq:MMDMatrix}. The smaller the conditional MMD is, the closer the conditional distributions of the source and the target datasets. Later, we will show how to minimize the above.

In order to estimate the conditional distribution of the target samples, the labels of the target samples need to be given. However the target labels are unknown in unsupervised domain adaptation. In such case, one can use some base classifiers, \emph{e.g.}, SVMs, MLPs, to obtain the pseudo labels for the target samples. In this work, the non-transfer neural networks are applied as the base classifier. And we iteratively update the labels of the target samples to ensure a good performance (the procedure is shown in Section \ref{sec:opt}). This empirically provides good performance.

\subsection{Final Objective Function}
By using the output of the last feature extraction layer as the input of the discrimination layer, we get DTN.
The feature extraction layers find a shared subspace, in which the marginal distributions of features of the source and the target datasets are matched. Then an adaptive classifier is trained in the new shared feature space to match the conditional distributions. The final objective of DTN is given in this part.

We use the negative log-likelihood, a commonly used loss function in the objective function.
It is  defined as:
\begin{equation}
\label{eq:Likelihood}
-\mathcal{L}(\mathbf{W}) = -\sum_{i = 1}^{n}\log{(P(y = y_i|\mathbf{x}_i,\mathbf{W}))}.
\end{equation}
where $\mathbf{W}$ is the projection matrix of the neural networks in Eq. \ref{eq:single_layer}. By integrating Eq. \ref{eq:MMD_Feature} and Eq. \ref{eq:MMD_Conditional} into Eq. \ref{eq:Likelihood}, we can get the final objective function of DTN:
\begin{equation}
\label{eq:objective_function}
\begin{aligned}
\mathbf{J}(\mathbf{W}) &=-\mathcal{L}(\mathbf{W})+\lambda \text{Tr}(\mathbf{H}(l-1)\mathbf{M}\mathbf{H}^T(l-1))\\
 &\quad +\mu\sum_{c = 1}^C \mathbf{q}_{c}^T\mathbf{M}\mathbf{q}_{c}\\
 & = -\mathcal{L}(\mathbf{W})+\lambda \MMD_{mar}+\mu \MMD_{con},
\end{aligned}
\end{equation}
where $\lambda$ and $\mu$ are marginal and condition distributions regulation parameters, which are further discussed in Section \ref{sec:exp}. Then, $\mathbf{W}$ can be calculated by minimizing Eq. \ref{eq:objective_function}. The final objective function does not depend on the exact form of the loss function and the structure of the network -- one can easily extend the formulation to other types of neural networks, such as CNNs and Deep Belief Networks (DBNs), in addition to MLPs.

\section{Optimization of Deep Transfer Network}
\label{sec:opt}
\subsection{The Optimization Procedure}
We first introduce some notations. For an arbitrary training sample $\mathbf{x}$, denote the posterior (conditional) probability output of the final classifier of this sample as $\mathbf{p}$. $\mathbf{p} \in \mathbb{R}^C$ is a $C$-dimensional vector, where each dimension represents a posterior probability of certain category. It is easy to show that: $\nabla_{\mathbf{h}(l-1)}(\MMD_{mar}) = $
\begin{equation}
\label{eq:MMD_Marginal_Derivative}
\left\{
        \begin{array}{r}
        2\frac{1}{n^s}(\frac{1}{n^s}\sum_{i = 1}^{n^s}\mathbf{h}^s_i(l-1)-\frac{1}{n^t}\sum_{j = 1}^{n^t}\mathbf{h}_j^t(l-1))\\
        \mathbf{x} \in D^{s}\\ \\
        -2\frac{1}{n^t}(\frac{1}{n^s}\sum_{i = 1}^{n^s}\mathbf{h}_i^s(l-1)-\frac{1}{n^t}\sum_{j = 1}^{n_t}\mathbf{h}_j^t(l-1))\\
        \mathbf{x} \in D^{t}
        \end{array}
        \right. ,
\end{equation}
and $\nabla_{\mathbf{p}}(\MMD_{con}) = $
\begin{equation}
\begin{split}
\label{eq:MMD_Conditional_Derivative}
		\left\{
        \begin{array}{cc}
        \frac{2}{n^s}(\frac{1}{n^s}\sum_{i = 1}^{n^s}\mathbf{p}^s_i-\frac{1}{n^t}\sum_{j = 1}^{n^t}\mathbf{p}^t_j) & \mathbf{x} \in D^{s}\\ \\
        -\frac{2}{n^t}(\frac{1}{n^s}\sum_{i = 1}^{n^s}\mathbf{p}^s_i-\frac{1}{n^t}\sum_{j = 1}^{n^t}\mathbf{p}^t_j) & \mathbf{x} \in D^{t}
        \end{array}
        \right.
\end{split}
\end{equation}

We then use the backpropagation algorithm to optimize the neural network. Denote one element in $\mathbf{W}$ as $W_{ij}$. The partial derivative for $W_{ij}$ of Eq. \ref{eq:objective_function} can be written as:
\begin{equation}
\label{eq:Final_Derivative}
\begin{aligned}
-\frac{\partial \mathcal{L}}{\partial W_{ij}}
 &+ \lambda(\nabla_{\mathbf{h}(l-1)}\MMD_{mar})^T(\frac{\partial \mathbf{h}(l-1)}{\partial W_{ij}})\\
 &+ \mu(\nabla_{\mathbf{p}}\MMD_{con})^T(\frac{\partial \mathbf{p}}{\partial W_{ij}}),
\end{aligned}
\end{equation}
where $\partial \mathbf{h}(l-1)/ \partial W_{ij}$ and $\partial \mathbf{p}/\partial W_{ij}$ are vectors consisting of partial derivatives of each element in $\mathbf{h}(l-1)$ and $\mathbf{p}$ with $W_{ij}$. They can be easily computed if the structure of the network is given. We can optimize the parameters of the network with stochastic gradient descent.

One issue to be addressed is that the computation of the marginal MMD or the conditional MMD requires taking all the source and the target samples into consideration. This is very inefficient, especially when the training dataset is large. Inspired by the idea of mini-batch stochastic gradient descent, we divide all the samples into different batches. Let $N$ be the number of the batches we want to build. The source and the target datasets are divided into $N$ parts. A training batch consist of $n^s/N$ source samples and $n^t/N$ target samples. It's easy to see:
\begin{equation}
\label{eq:MMD_MiniBatch}
\begin{aligned}
&\MMD = \parallel \frac{1}{n^s}\sum_{i = 1}^{n^s}\mathbf{x}^s_i- \frac{1}{n^t}\sum_{j = 1}^{n^t}\mathbf{x}^t_j \parallel_2^2 \\
&\leqslant N\sum_{k = 1}^{N} \parallel \frac{1}{n^s}\sum_{\mathbf{x}^s_i \in \mathcal{B}_k}\mathbf{x}^s_i - \frac{1}{n^t}\sum_{\mathbf{x}^t_j \in \mathcal{B}_k}\mathbf{x}^t_j \parallel_2^2\\
& = N^2\sum_{k = 1}^{N} \parallel \frac{1}{n_k^s}\sum_{\mathbf{x}^s_i \in \mathcal{B}_k}\mathbf{x}^s_i - \frac{1}{n_k^t}\sum_{\mathbf{x}^t_j \in \mathcal{B}_k}\mathbf{x}^t_j \parallel_2^2,
\end{aligned}
\end{equation}
where $\mathcal{B}$ is the set of all the batches, $\mathcal{B}_k$ is the $k~$th batch in $\mathcal{B}$. $n_k^s = n^s/N$ and $n_k^t = n^t/N$ are the numbers of the source and the target datasets in $\mathcal{B}_k$. Therefore, if all the mini-batches are matched very well, the whole datasets are matched.
Instead of computing MMD over the whole datasets, we compute MMD over every single batch.

\begin{figure}
\centering
    \includegraphics[width=0.3\linewidth]{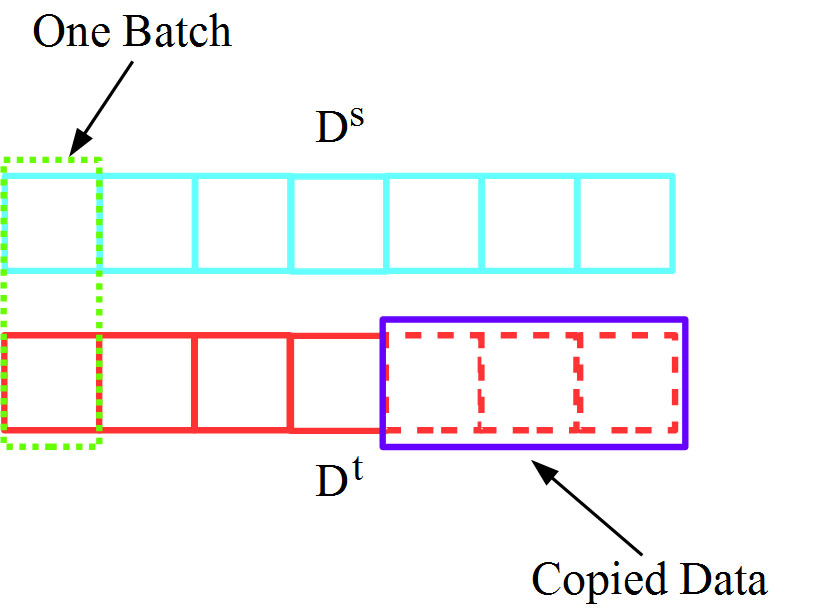}
   \caption{Method of building the batches of data. The samples in $D^t$ are copied (dashed rectangles) so that the two datasets have same sizes. The training batches are then built by randomly picking samples from two datasets.}
   \label{fig:Mini_Batch}
\end{figure}

The training batches are built as follows. First, the samples of the smaller dataset are randomly picked and copied such that the source and the target datasets have the same number of samples. Assume the size of single batch is $S$. Then batches are filled by randomly picking $S/2$ samples from the source dataset and other $S/2$ samples from the target dataset, until all the samples are selected. The batch size $S$ should be big enough so that every mini-batch can reflect the variance of the whole dataset. An empirical analysis on $S$ is given in section \ref{sec:exp}. Figure \ref{fig:Mini_Batch} illustrates the process of building the batches of data. Finally, gradient descent on the mini-batch is applied to optimize the objective function.

As discussed in Section \ref{sec:dtn}, the outputs of a non-transfer neural networks are utilized as the pseudo labels of the target samples. It is not surprising that the more accurate the labels of the target data are provided, the better performance of the final classifier. In particular, DTN can use its output as input to improve itself. We find that iteratively updating the labels of the target samples during training can significantly improve the performance of DTN, especially on large-scale datasets. We analyze the convergence property of DTN in Section \ref{sec:exp}. Algorithm \ref{alg:DTN} summarises the optimization of DTN.
\begin{algorithm}[tb]%
\caption{Optimization of the Deep Transfer Network}
\begin{algorithmic}[1]
\REQUIRE
Source data with label $\mathbf{X}^s, Y^s$ and target data $\mathbf{X}^t$.\\
\ENSURE
 Parameters of deep transfer network $\mathbf{W}$ and predicted labels of the target samples $\bar{Y}^t$.\\
\PRINT
  \STATE Set $i = 0$. Get $\bar{Y}^t_0$ by training a baseline neural network with $\mathbf{X}^s, Y^s$ and testing with $\mathbf{X}^t$.
  \REPEAT
  \STATE $i = i+1$.
  \STATE Make mini-batches.
  \STATE Get $\mathbf{W}$ by optimizing Eq.\ref{eq:objective_function} using source data $\mathbf{X}^s, Y^s$ and target data $\mathbf{X}^t, \bar{Y}_{i-1}^t$.
  \STATE Predict $\mathbf{X}^t$ with the network to get $\bar{Y}^t_i$.
  \UNTIL $\bar{Y}^t_{i} = \bar{Y}^t_{i-1}$ \OR $i>T$.
\RETURN $\mathbf{W}$, $\bar{Y}^t = \bar{Y}^t_{i}$.
\end{algorithmic}
\label{alg:DTN}
\end{algorithm}

\subsection{Computational Complexity}
Let $n = \max(n_s,n_t)$ be the size of the bigger dataset, and the number of batches in $\mathcal{B}$ to be $N$. For each mini-batch, the cost of computing Eq. \ref{eq:MMD_Marginal_Derivative} and \ref{eq:MMD_Conditional_Derivative} are both $\mathcal{O}(S)$, where $S$ is the size of one batch. More samples in a single batch always leads to higher computational cost when computing the gradient over the batch.
The computational complexity of forward propagation over one batch is also $\mathcal{O}(S)$. Since there are a total of $N$ batches and $S \times N = 2n$, the total computational complexity of the backpropagation in DTN over the whole datasets is $\mathcal{O}(n)$.
Unlike other high order domain adaptation algorithms, such as, Transfer Joint Matching ($\mathcal{O}(n^2)$) \cite{long2014transfer} and Adaptation Regularization based Regularized Least Squares ($\mathcal{O}(n^3)$) \cite{long2014adaptation}, the execution time of DTN grows linearly in terms of the number of input samples. Thus, DTN is suitable to be applied on large-scale datasets.

\section{Experiments}
\label{sec:exp}
In this section, we conduct comprehensive experiments to evaluate the performance of DTN. Besides standard domain adaptation datasets widely used in existed literatures, we also develop two new settings which have 10 times more samples than existed datasets.

\begin{table*}[t]%
\caption{Classification Accuracies (\%) on Office-Caltech Dataset}
\label{ClassificationAccuracy_Surf}
\begin{center}
\begin{tabular}{c|ccccccccccccccccc}
\hline
& \multicolumn{13}{c}{Database}\\
Method &A/W  &A/D  &A/C  &W/A  &W/D  &W/C  &D/A  &D/W  &D/C  &C/A  &C/W  &C/D  &Avg\\
\hline
NN     &29.83&25.48&26.00&22.96&59.24&19.86&28.50&63.39&26.27&23.70&25.76&25.48&31.37\\
FSSL   &34.35&26.37&33.91&29.53&76.79&25.85&30.61&74.99&27.89&35.88&32.32&37.53&38.84\\
TCA    &35.25&34.39&40.07&28.81&85.99&29.92&31.42&86.44&32.06&45.82&30.51&35.67&43.03\\
GFK    &41.69&37.58&37.58&27.77&82.17&30.10&33.61&79.66&30.45&41.54&35.93&43.31&43.45\\
TJM    &42.03&45.22&39.45&29.96&\textbf{89.17}&30.19&32.78&85.42&31.43&46.76&38.98&44.59&46.33\\
RLS    &38.98&38.85&42.20&38.62&78.34&34.46&34.13&79.66&\textbf{33.21}&53.86&49.49&44.59&47.20\\
ARRLS  &38.98&38.85&42.48&\textbf{39.87}&78.34&\textbf{34.73}&\textbf{37.16}&82.71&32.24&\textbf{54.91}&50.51&44.59&47.95\\
MLP    &36.80&46.00&41.45&33.26&70.00&32.91&31.37&74.40&29.64&51.05&48.80&45.33&45.08\\
DTN    &\textbf{43.00}&\textbf{56.00}&\textbf{42.90}&36.89&84.00&34.18&34.89&\textbf{87.50}&32.27&54.00&\textbf{58.50}&\textbf{56.00}&\textbf{51.68}\\
\hline
\end{tabular}%
\end{center}
\end{table*}

\subsection{Domain Adaptation on Small-Scale Datasets}

We use the publicly available Office-Caltech dataset\footnote{From https://www.eecs.berkeley.edu/\~{}jhoffman/domainadapt/} to evaluate domain adaptation algorithms on small-scale dataset. Office-Caltech dataset, first released by Gong \etal. \cite{gong2012geodesic}, consists of the Office dataset and the Caltech-256 dataset. Office dataset contains three domains: Amazon (online merchants images downloaded from www.amazon.com), DSLR (image of corresponding merchants captured by DSLR camera in realistic environments) and Webcam (image captured by webcam). And Caltech-256 dataset has 30,607 images in 256 categories. The four domains (Amazon (A), Webcam (W), DSLR (D), and Caltech (C)) share 10 object categories in total. In the corresponding categories they have 958, 295, 157 and 1,123 image samples respectively, with a total of 2,533 images. For all the images, SURF features are extracted and quantized into an 800-bin histogram with codebook trained from a subset of images in Amazon. The dataset is a standard benchmark for evaluating domain adaptation algorithms. By randomly selecting two different domains out of four domains to form the source and the target datasets, we can have $12$ source/target cross-domain pairs, marked as A/W, A/D, A/C,$\ldots$, C/D.

We report results of DTN on all the 12 pairs and compared to the methods reported in \cite{long2014adaptation} and \cite{long2014transfer}. The methods include:
\begin{itemize}
\setlength{\itemsep}{-\itemsep}
    \item 1-Nearest Neighbor Classifier (NN)
    \item Joint Feature Selection and Subspace Learning (FSSL) \cite{gu2011joint} + NN
    \item Transfer Component Analysis (TCA)  \cite{pan2011domain} + NN
    \item Geodesic Flow Kernel (GFK) \cite{gong2012geodesic} + NN
    \item Transfer Joint Matching (TJM) \cite{long2014transfer} + NN
    \item Regularized Least Squares (RLS)
    \item Adaptation Regularization based Regularized Least Squares (ARRLS) \cite{long2014adaptation}
    \item Multiple Layer Perceptrons (MLP)
\end{itemize}
Among those, NN, MLP and RLS are the base classifiers without knowledge transfer. FSSL, TCA, GFK, TJM are subspace learning methods, while ARRLS is classifier transduction method. In particular, MLP is the non-transfer base classifier of DTN. And ARRLS is the method most closely related to DTN, since they both use marginal MMD and conditional MMD to evaluate the discrepancy between distributions. However, their underlying classification models are different. DTN is implemented using Theano, a parallel Python package \cite{bergstra+al:2010-scipy}.

We follow the same protocol in \cite{Long_2013_CVPR,long2014transfer} in the evaluation.
The performance of DTN depends on the architecture of the neural networks.
Besides the architecture of the neural networks, DTN involves four meta parameters: the regularization parameter for marginal distribution matching $\lambda$, the regularization parameter for conditional distribution matching $\mu$, the batch size $S$ which determines the size of the mini-batch, and the iteration number $T$ controlling how many times the labels of the target samples are updated during the training process. For small-scale datasets, we only use the labels by non-transfer MLP during training.

In this experiment, we only use a neural network with one hidden layer. In other words, the base neural network only have one feature extraction layer and one discrimination layer. The number of hidden nodes is 500. We use the default setting $\lambda = \mu = 10$ throughout the whole dataset. We empirically set batch size $S = 200$, which means that every mini-batch consists of 100 source samples and 100 target samples. However, since samples in DSLR and Webcam domains are too few to build enough batches for training, we set $s = 100$ for all the source/target pairs which involve DSLR or/and Webcam domains. The classification accuracy is used as an evaluate metric as in \cite{Long_2013_CVPR,long2014transfer}.

Table \ref{ClassificationAccuracy_Surf} shows all the obtained classification accuracies on all the source/target pairs in Office-Caltech dataset for DTN and all the eight baseline methods. We observe that, DTN achieves the best performance on 6 out of 12 datasets, and on the other two datasets, the performance of DTN is only slightly worse (less than 1\% of the classification accuracy) than the best baseline method. The average classification accuracy on all the 12 datasets of DTN is 51.68\%, which is 3.73\% higher compared to the best baseline method ARRLS. Compared to its non-transfer base classifier MLP, DTN gains improvement on all 12 datasets and finally outperforms MLP by 6.60\% on average accuracy. This fact proves that, simultaneously minimizing MMD of both the marginal and the conditional distributions of the source and the target datasets can significantly improve the performance of the trained classifier on the unlabeled target dataset.

We also notice DTN performs better than ARRLS. These two methods both use MMD to evaluate the marginal and the conditional distributions divergences between two datasets. The only difference is that the base classifiers, DTN uses MLP and ARRLS uses RLS. This proves that the combination of the best paradigms of both worlds (neural networks in recognition, and matching both the marginal and the conditional distributions in domain adaptation) works well. We also observe that methods matching both the marginal and the conditional distributions (TJM, ARRLS and DTN) always perform better than methods matching marginal distributions only (TCA and GFK). The reason is that, both the marginal and the conditional distributions may change over different domains in real-world problems. Matching marginal distributions only cannot guarantee small discrepancy between conditional distributions, and the discriminative directions of the source and the target domains may still be different \cite{long2014adaptation}.

\begin{table*}[t]%
\caption{Classification Accuracies (\%) on Large-Scale Benchmark Datasets}
\label{ClassificationAccuracyDigit}
\begin{center}
\begin{tabular}{c|ccccc}
\hline
          & \multicolumn{5}{c}{Method}\\
Database  &GFK    &RLS    &ARRLS  &MLP (CNN)     &DTN (ours)\\
\hline
USPS/MNIST&33.71&16.57&52.09&44.47&\textbf{81.04}\\
CIFAR/VOC &49.15&62.43&71.73&59.04&\textbf{73.60}\\
\hline
\end{tabular}%
\end{center}
\end{table*}

\subsection{Domain Adaptation on Large-Scale Datasets}
We develop two new large-scale settings to evaluate the performance of domain adaptation algorithms, which are USPS/MNIST dataset and CIFAR/VOC dataset. These two datasets are almost 10 times larger than current datasets. For both the settings, we use one dataset as the source set and the other one as the target set to evaluate our method.

\textbf{USPS/MNIST.} USPS\footnote{From http://www.csie.ntu.edu.tw/\~{}cjlin/libsvmtools/\\datasets/multiclass.html$\sharp$usps} and MNIST\footnote{From http://yann.lecun.com/exdb/mnist/} datasets are two handwriting datasets widely used in classification algorithm evaluation. USPS dataset consists of 10,000 training and testing images with image size of 16$\times$16, while MNIST dataset has 50,000 training images with image size of 28$\times$28. We follow the same preprocessing procedure in \cite{Long_2013_CVPR}. However, rather than randomly sampling a subset, we use all the training and testing images in the USPS dataset as the source samples and all the training samples in the MNIST dataset as the target samples. Thus, our USPS/MNIST dataset has 60,000 samples in total, which is 20 times larger than the one used in \cite{Long_2013_CVPR}.

\begin{figure}
\begin{center}
\subfigure[Sample images in CIFAR-10 dataset.]{\label{fig:cifar} \includegraphics[width=0.6\linewidth]{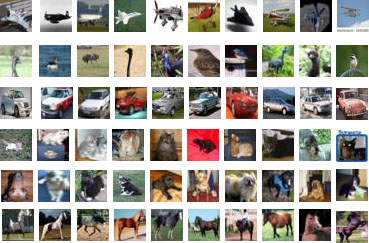}}
\subfigure[Sample images in Pascal VOC 2012 dataset.]{\label{fig:voc}\includegraphics[width=0.8\linewidth]{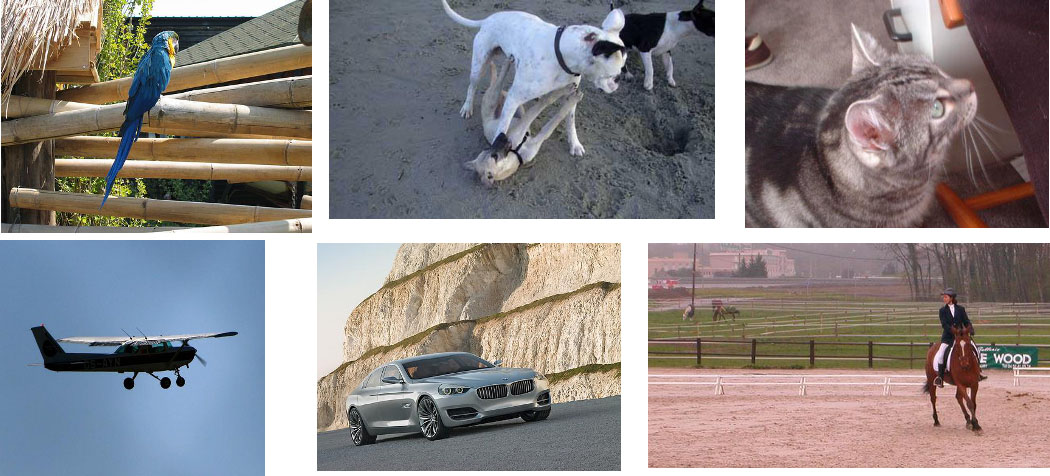}}
\end{center}
\caption{Sample images in CIFAR/VOC dataset.}
\label{fig:sample_images}
\end{figure}

\textbf{CIFAR/VOC.} CIFAR-10 dataset\footnote{From http://www.cs.toronto.edu/\~{}kriz/cifar.html} \cite{krizhevsky2009learning} is a labeled subset of the 80 million tiny images dataset \cite{torralba200880}. It consists of 60,000 32$\times$32 colour images in 10 categories, 6,000 images per class. Pascal VOC 2012 dataset\footnote{From http://pascallin.ecs.soton.ac.uk/challenges/VOC/voc2012/} \cite{pascal-voc-2012} is designed for recognizing objects in realistic scenes. It has 20 classes with a total of 11,530 images. Some sample images of CIFAR-10 and Pascal VOC 2012 are shown in figure \ref{fig:sample_images}. The CIFAR-10 dataset consists of very tiny images while the images in Pascal VOC 2012 look like images taken from the Internet. They follow very different distributions. CIFAR-10 dataset and Pascal VOC 2012 dataset share 6 semantic categories, ``aeroplane/airplane'', ``bird'', ``car'', ``cat'', ``dog'' and ``horse''. We therefore construct CIFAR/VOC dataset by randomly sampling 15,000 images (2,500 per category) from CIFAR-10 dataset to form the source dataset and selecting all the samples in Pascal VOC 2012 dataset to form the target dataset. For all the 17,720 image samples, DeCAF feature \cite{jia2014caffe} is extracted. We use the 6th layer output, which leads to a 4096-dimension feature.

\begin{figure*}[!htb]
\begin{center}
\subfigure[Distribution Regularization]{\label{fig:Distribution_Sensitivity} \includegraphics[width=0.242\linewidth]{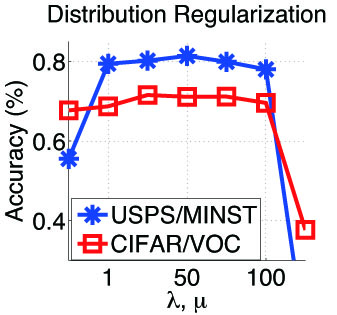}}
\subfigure[Batch Size]{\label{fig:Batch_Size}\includegraphics[width=0.242\linewidth]{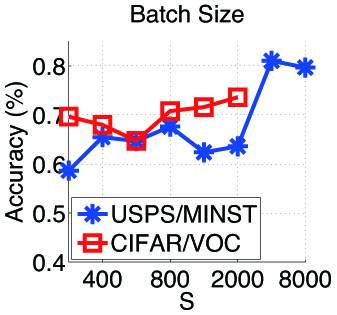}}
\subfigure[Iteration]{\label{fig:Iteration_Number} \includegraphics[width=0.242\linewidth]{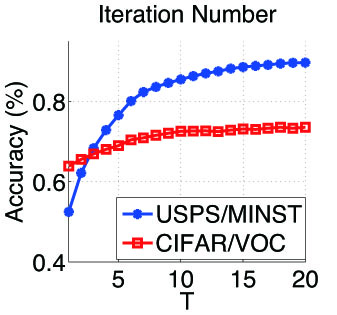}}
\subfigure[Execution Time]{\label{fig:Execution_Time} \includegraphics[width=0.242\linewidth]{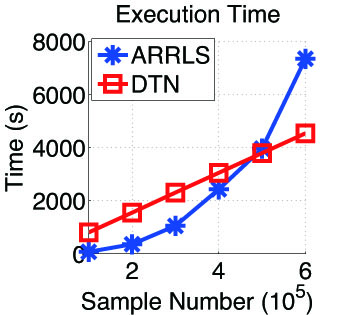}}
\end{center}
\caption{Parameters analysis on USPS/MNIST and CIFAR/VOC datasets and execution time on various sizes USPS/MNIST datasets.}
\label{fig:Parameter_Sensitivity}
\end{figure*}

A convolutional neural network (CNN) is applied to USPS/MNIST dataset as the base classifier. The CNN is based on LeNet model, which was first proposed in \cite{lecun1998gradient}. It has two convolutional and max-pooling layers with filter size 3$\times$3 and a fully-connected layer with 500 nodes. A multilayer perceptron with 3 hidden layers is used as the base classifier for CIFAR/VOC dataset. The numbers of the nodes in each hidden layer are 2,000, 1,000 and 500 respectively. For parameter settings, we also set the distribution regulation parameters to the default setting $\lambda = \mu = 10$. The mini-batch size $S$ for MINST/USPS is set to 4,000, which means that one mini-batch consists of 2,000 source samples and 2,000 target samples. For CIFAR/VOC, $S = 2,000$. Iteration number $T$ is set to 20, meaning that the labels of the target samples are iteratively updated 20 times during training.

Table \ref{ClassificationAccuracyDigit} shows the results of DTN and some baseline methods. We observe that the classification accuracy of DTN on USPS/MNIST and CIFAR/VOC datasets are 81.04\% and 73.60\% respectively. Compared to its non-transfer base classifier neural networks, it achieves improvements of 36.57\% and 14.56\%. The best baseline method is ARRLS, which gets accuracies of 52.09\% and 71.73\% and gains 35.52\% and 9.3\% improvements from its non-transfer base classifier RLS. Compared to the best baseline method (ARRLS), DTN gains improvements of 28.95\% and 1.87\%. We observe that DTN performs significantly better than any other baseline method on USPS/MNIST dataset. It's because of the extraordinary performance of its base classifier CNN in digit recognition.

Table \ref{ExecuteTime} shows the execution time of some algorithms run on the large-scale datasets. DTN has competitive speed compared to the baseline method with best performance (ARRLS). In order to directly display the computational complexity, we randomly sample USPS/MNIST to different sizes, from 10,000 to 60,000 and plot the execution time of DTN and ARRLS on Figure \ref{fig:Execution_Time}. It shows the execution time of DTN grows almost linearly with the number of the training samples, however, ARRLS grows much faster. It should also be remarked here that ARRLS requires over 100GB memory to save the kernel matrix, while less than 3GB video memory is enough for DTN. The experimental result proves that DTN can deal with very large-scale dataset with appealing transfer performance.

\subsection{Parameter Sensitivity}
We conduct parameter sensitivity experiment on USPS/MNIST and CIFAR/VOC dataset. Distribution matching parameters $\lambda$, $\mu$, size of the mini-batch $S$ and iteration number $T$ are evaluated.

\begin{table}[t]%
\caption{Execution Time (s) on Large-Scale Benchmark Datasets}
\label{ExecuteTime}
\begin{center}
\begin{tabular}{c|ccc}
\hline
          & \multicolumn{3}{c}{Method}\\
Database  &GFK&ARRLS&DTN (ours)\\
\hline
CIFAR/VOC &153&640 &1,428\\
USPS/MNIST&34 &7,346&4,548\\
\hline
\end{tabular}%
\end{center}
\end{table}

\textbf{Distribution Regularization Parameters $\lambda$ $,$ $\mu$}.  $\lambda$ controls the level of marginal distribution matching, and $\mu$ controls the level of conditional distribution matching. The larger the values are, the smaller the differences of distributions. For simplification, we set $\lambda = \mu$. Figure \ref{fig:Distribution_Sensitivity} shows the classification accuracies with different values taken from $\{0,1,5,10,50,100,500\}$. We noticed that $\lambda$ and $\mu$ can be chosen from $[1,100]$. Throughout the whole experiment part, we choose $\lambda = \mu = 10$ as default.

\textbf{Batch Size $S$}. The mini-batch is the basic unit for evaluating the distribution of the database and optimizing the objective function. The size $S$ should be large enough to contain enough samples in the batch so that it can reflect the distribution of the whole dataset. Figure \ref{fig:Batch_Size} shows the classification accuracies with different values taken from $\{200,400,600,800,1000,$ $2000,4000,8000\}$. We notice that bigger batch always leads to better performance. For large-scale dataset, $S$ can be chosen from $[2000,8000]$. On the other hand, the size of the batch should not to be too large, due to limited GPU memory (experiments on CIFAR/VOC with batch size 4,000 and 8,000 failed due to insufficient video memory). Finally, we choose $S = 4,000$ for USPS/MNIST and $S = 2,000$ for CIFAR/VOC.

\textbf{Iteration Number $T$}. DTN can use its output as its input and alternatively improve the performance. $T$ determines how many times the labels of the target datasets are updated during training. Figure \ref{fig:Iteration_Number} shows that the classification accuracy increases steadily with more iterations and finally converges (the classification accuracies of USPS/MNIST are plotted every 4 iterations). Finally, we choose $T = 20$ for experiments on large-scale datasets.

The unsupervised DTN can also be extended to a semi-supervised version easily. Labeled samples in target dataset can always provide better estimation of the distribution of the target data.

\section{Conclusion}
We proposed the Deep Transfer Network (DTN), which combines the best paradigms in object recognition (neural network) and domain adaptation (matching both the marginal and the conditional distributions). The structure of neural networks makes efficient modeling and optimization of the distribution matching process.

DTN has a computational complexity $\mathcal{O}(n)$, making it suitable to be applied on large-scale domain adaptation problems. Comprehensive experiments showed that DTN is effective on a variety of benchmark datasets and it significantly outperforms competitive methods especially on large-scale problems.


\bibliographystyle{ieee}
\bibliography{egbib.bib}
\end{document}